\documentclass[runningheads]{llncs}

\usepackage{graphicx}
\usepackage{amsfonts}
\usepackage{physics}
\usepackage{multirow}
\usepackage{enumitem}
\usepackage{tablefootnote}
\usepackage{amsmath}

\begin{document}

\title{Fair Outlier Detection}

\author{Deepak P\inst{1,2} \and
Savitha Sam Abraham\inst{2}}
\authorrunning{P, Deepak and Sam Abraham, Savitha}
\institute{Queen's University Belfast, UK \and
Indian Institute of Technology Madras, India \\
\email{deepaksp@acm.org \ \ savithas@cse.iitm.ac.in}}

\maketitle 

\begin{abstract}
An outlier detection method may be considered fair over specified sensitive attributes if the results of outlier detection are not skewed towards particular groups defined on such sensitive attributes. In this paper, we consider, for the first time to our best knowledge, the task of fair outlier detection. Our focus is on the task of fair outlier detection over multiple multi-valued sensitive attributes (e.g., gender, race, religion, nationality, marital status etc.), one that has broad applications across web data scenarios. We propose a fair outlier detection method, {\it FairLOF}, that is inspired by the popular {\it LOF} formulation for neighborhood-based outlier detection. We outline ways in which unfairness could be induced within {\it LOF} and develop three heuristic principles to enhance fairness, which form the basis of the {\it FairLOF} method. Being a novel task, we develop an evaluation framework for fair outlier detection, and use that to benchmark {\it FairLOF} on quality and fairness of results. Through an extensive empirical evaluation over real-world datasets, we illustrate that {\it FairLOF} is able to achieve significant improvements in fairness at sometimes marginal degradations on result quality as measured against the fairness-agnostic {\it LOF} method. 
\end{abstract}

\section{Introduction}

There has been much recent interest in incorporating fairness constructs into data analytics algorithms, within the broader theme of algorithmic fairness~\cite{10.1145/3376898}. The importance of fairness in particular, and democratic values in general, cannot be overemphasized in this age when data science algorithms are being used in very diverse scenarios to aid decision making that could affect lives significantly. The vast majority of fair machine learning work has focused on supervised learning, especially on classification (e.g.,~\cite{huang2019stable,zafar2015fairness}). There has also been some recent interest in ensuring fairness within unsupervised learning tasks such as clustering~\cite{DBLP:conf/edbt/Abraham0S20}, retrieval~\cite{zehlike2017fa} and recommendations~\cite{patro2019incremental}. In this paper, we explore the task of fairness in outlier detection, an analytics task of wide applicability in myriad scenarios. To our best knowledge, this is the first work on embedding fairness within outlier detection. The only work so far on fairness and outlier detection~\cite{fairod2020} focuses on analyzing outlier detection algorithms for fairness, a significantly different task, for which a human-in-the-loop method is proposed. 

\noindent{\bf Outlier Detection and Fairness:} The task of outlier detection targets to identify {\it deviant} observations from a dataset, and is usually modelled as an unsupervised task; ~\cite{chandola2007outlier} provides a review of outlier detection methods. The classical outlier characterization, due to Hawkins~\cite{hawkins1980identification}, considers outliers as 'observations that deviate so much from other observations as to arouse suspicion that they were generated by a different process'. Applications of outlier detection range across varied application domains such as network intrusions~\cite{jabez2015intrusion}, financial fraud~\cite{pawar2014survey} and medical abnormalities~\cite{kumar2008outlier}. Identification of non-mainstream behavior, the high-level task that outlier detection accomplishes, has a number of applications in new age data scenarios. Immigration checks at airports might want to carry out detailed checks on {\it `suspicious'} people, while AI is likely used in proactive policing to identify {\it `suspicious'} people for stop-and-frisk checks. In this age of pervasive digitization, {\it `abnormality'} in health, income or mobility patterns may invite proactive checks from healthcare, taxation and policing agencies. Identification of such {\it abnormal} and {\it suspicious} patterns are inevitably within the remit of outlier detection. The nature of the task of outlier detection task makes it very critical when viewed from the perspective of fairness. Even if information about ethnicity, gender, religion and nationality be hidden (they are often not hidden, and neither is it required to be hidden under most legal regulations) from the database prior to outlier identification, information about these attributes are likely inherently spread across other attributes. For example, geo-location, income and choice of professions may be correlated with ethnic, gender, religious and other identities. The identification of non-mainstream either falls out from, or entails, an analogous and implicit modelling of mainstream characteristics in the dataset. The mainstream behavior, by its very design, risks being correlated with majoritarian identities, leading to the possibility of minority groups being picked out as outliers significantly more often. Interestingly, there have been patterns of racial prejudice in such  settings\footnote{https://www.nyclu.org/en/stop-and-frisk-data}. 


\noindent{\bf Outlier Detection and the Web:} Web has emerged, over the past decades, as a rich source of unlabelled digital data. Thus, the web likely presents the largest set of scenarios involving outlier detection. Each user on the web leaves different cross-sections of digital footprints in different services she uses, together encompassing virtually every realm of activity; this goes well beyond the public sector applications referenced above. In a number of scenarios, identified as an outlier could lead to undesirable outcomes for individuals. For example, mobility outliers may receive a higher car insurance quote, and social media outliers may be subjected to higher scrutiny (e.g., Facebook moderation). It is important to ensure that such undesirable outcomes be distributed fairly across groups defined on protected attributes (e.g., gender, race, nationality, religion etc.) for ethical reasons and to avoid bad press\footnote{https://www.cnet.com/features/is-facebook-censoring-conservatives-or-is-moderating-just-too-hard/}. 

\noindent{\bf Our Contributions:} We now outline our contributions in this paper. First, we characterize the task of fair outlier detection under the normative principle of disparate impact avoidance~\cite{barocas2016big} that has recently been used in other unsupervised learning tasks~\cite{chierichetti2017fair,DBLP:conf/edbt/Abraham0S20}. Second, we develop a fair outlier detection method, {\it FairLOF}, based on the framework of {\it LOF}~\cite{breunig2000lof}, arguably the most popular outlier detection method. Our method is capable of handling multiple multi-valued protected attributes, making it feasible to use in sophisticated real-world scenarios where fairness is required over a number of facets. Third, we outline an evaluation framework for fair outlier detection methods, outlining quality and fairness metrics, and trade-offs among them. Lastly, through an extensive empirical evaluation over real-world datasets, we establish the effectiveness of {\it FairLOF} in achieving high levels of fairness at small degradations to outlier detection quality. 



\section{Related Work}

Given the absence of prior work on fair outlier detection methods, we cover related work across outlier detection and fairness in unsupervised learning. 

\noindent{\bf Outlier Detection Methods:} Since obtaining labelled data containing outliers is often hard, outlier detection is typically modelled as an unsupervised learning task where an unlabelled dataset is analyzed to identify outliers within it. That said, supervised and semi-supervised approaches do exist~\cite{chandola2007outlier}. We address the unsupervised setting. The large majority of work in unsupervised outlier detection may be classified into one of two families. The first family, that of {\it global methods}, build a dataset-level model, and regard objects that do not conform well to the model, as outliers. The model could be a clustering~\cite{yu2002findout}, Dirichlet mixture~\cite{fan2011unsupervised} or others~\cite{domingues2018comparative}. Recent research has also explored the usage of auto-encoders as a global model, the reconstruction error of individual data objects serving as an indication of their outlierness; {\it RandNet}~\cite{chen2017outlier} generalizes this notion to determine outliers using an ensemble of auto-encoders. The second family, arguably the more popular one, is that of {\it local methods}, where each data object's outlierness is determined using just its neighborhood within a relevant similarity space, which may form a small subset of the whole dataset. The basic idea is that the outliers will have a local neighborhood that differs sharply in terms of characteristics from the extended neighborhood just beyond. {\it LOF}~\cite{breunig2000lof} operationalizes this notion by quantifying the contrast between an object's local density (called local reachability density, as we will see) and that of other objects in its neighborhood. Since the {\it LOF} proposal, there has been much research into local outliers, leading to work such as {\it SLOM}~\cite{chawla2006slom}, {\it LoOP}~\cite{kriegel2009loop} and {\it LDOF}~\cite{zhang2009new}. Schubert et al~\cite{schubert2014local} provide an excellent review of local outlier detection, including a generalized three phase meta-algorithm that most local outlier detection methods can be seen to fit in. Despite much research over the last two decades, {\it LOF} remains the dominant method for outlier detection, continuously inspiring systems work on making it efficient for usage in real-world settings (e.g.,~\cite{babaei2019detecting}). Accordingly, the framework of {\it LOF} inspires the construction of {\it FairLOF}. 

\noindent{\bf Fairness in Unsupervised Learning:} There has been much recent work on developing fair algorithms for unsupervised learning tasks such as clustering, representation learning and retrieval. Two streams of fairness are broadly used; {\it group fairness} that targets to ensure that the outputs are fairly distributed across groups defined on sensitive attributes, and {\it individual fairness} which strives to limit possibilities of similar objects receiving dissimilar outcomes. Individual fairness is typically agnostic to the notion of sensitive attributes. Our focus, in this paper, is on group fairness in outlier detection. For group fairness in clustering, techniques differ in where they embed the fairness constructs; it could be at the pre-processing step~\cite{chierichetti2017fair}, within the optimization framework~\cite{DBLP:conf/edbt/Abraham0S20} or as a post-processing step to re-configure the outputs~\cite{DBLP:conf/nips/BeraCFN19}. {\it FairPCA}~\cite{olfat2019convex}, a fair representation learner, targets to ensure that objects are indistinguishable wrt their sensitive attribute values in the learnt space. Fair retrieval methods often implement group fairness as parity across sensitive groups in the top-$k$ outputs~\cite{asudeh2019designing}. The techniques above also differ in another critical dimension; the number of sensitive attributes they can accommodate. Some can only accommodate one binary sensitive attribute, whereas others target to cater to fairness over multiple multi-valued sensitive attributes; a categorization of clustering methods along these lines appears in~\cite{DBLP:conf/edbt/Abraham0S20}. 

In contrast to such work above, there has been no exploration into fair outlier detection. The only related effort in this space so far, to our best knowledge, is that of developing a human-in-the-loop decision procedure to determine whether the outputs of an outlier detection is fair~\cite{fairod2020}. This focuses on deriving {\it explanations based on sensitive attributes} to distinguish the outputs of an outlier detection method from the {\it `normal'} group. If no satisfactory explanation can be achieved, the black-box outlier detection method can be considered fair. The human is expected to have domain knowledge of the task and data scenario to determine parameters to identify what is unfair, and interpret explanations to judge whether it is indeed a case of unfairness. 

\section{Problem Definition}

\noindent{\bf Task Setting:} Consider a dataset $\mathcal{X} = \{ \ldots, X, \ldots\}$ and an object pairwise distance function $d: \mathcal{X} \times \mathcal{X} \rightarrow \mathbb{R}$ that is deemed relevant to the outlier detection scenario. Further, each data object is associated with a set of sensitive attributes $\mathcal{S} = \{ \ldots, S, \ldots\}$ (e.g., gender, race, nationality, religion etc.) which are categorical and potentially multi-valued, $V(S)$ being the set of values that a sensitive attribute, $S$, can take. $X.S \in V(S)$ indicates the value assumed by object $X$ for the sensitive attribute $S$. Thus, each multi-valued attribute $S$ defines a partitioning of the dataset into $|V(S)|$ parts, each of which comprise objects that take the same distinct value for $S$. 

\noindent{\bf Outlier Detection:} The task of (vanilla) outlier detection is that of identifying a small subset of objects from $\mathcal{X}$, denoted as $\mathcal{O}$, that are deemed to be outliers. Within the {\it local outliers} definition we adhere to, it is expected that objects in $\mathcal{O}$ differ significantly in local neighborhood density when compared to other objects in their neighborhoods. In typical scenarios, it is also expected that $|\mathcal{O}| = t$, where $t$ is a pre-specified parameter. The choice of $t$ may be both influenced by the dataset size (e.g., $t$ as a fixed fraction of $|\mathcal{X}|$) and/or guided by practical considerations (e.g., manual labour budgeted to examine outliers). 

\noindent{\bf Fair Outlier Detection:} The task of fair outlier detection, in addition to identifying outliers, considers ensuring that the distribution of sensitive attribute groups among $\mathcal{O}$ reflect that in $\mathcal{X}$ as much as possible. This notion, referred to interchangeably as representational parity or disparate impact avoidance, has been the cornerstone of all major fair clustering algorithms (e.g.,~\cite{chierichetti2017fair,DBLP:conf/edbt/Abraham0S20,DBLP:conf/nips/BeraCFN19}), and is thus a natural first choice as a normative principle for fair outlier detection. As a concrete example, if gender is a sensitive attribute in $\mathcal{S}$, we would expect the gender ratio within $\mathcal{O}$ to be very close to, if not exactly equal to, the gender ratio in $\mathcal{X}$. Note that fairness is complementary and often contradictory to ensuring that the top neighborhood-outliers find their place in $\mathcal{O}$; the latter being the only consideration in (vanilla) outlier detection. Thus, fair outlier detection methods such as {\it FairLOF} we develop, much like fair clustering methods, would be evaluated on two sets of metrics: (a) {\it 'Quality'} metrics that measure how well objects with distinct local neighborhoods are placed in $\mathcal{O}$, and (b) {\it Fairness} metrics that measure how well they ensure that the dataset-distribution of sensitive attribute values are preserved within $\mathcal{O}$. We will outline a detailed evaluation framework in a subsequent section. Good fair outlier detection methods would be expected to achieve good fairness while suffering only small degradations in quality when compared against their vanilla outlier detection counterparts. 

\noindent{\bf Motivation for Representational Parity:} It may be argued that the distribution of sensitive attribute groups could be legitimately different from that in the dataset. For example, one might argue that outlying social media profiles that correlate with crime may be legitimately skewed towards certain ethnicities since propensity for crimes could be higher for certain ethnicities than others. The notion of representational parity disregards such assumptions of skewed apriori distributions, and seeks to ensure that the inconvenience of being classed as an outlier be shared proportionally across sensitive attribute groups. This argument is compelling within scenarios of using outlier detection in databases encompassing information about humans. In particular, this has its roots in the distributive justice theory of {\it luck egalitarianism}~\cite{knight2009luck} that distributive shares be not influenced by arbitrary factors, especially those of {\it `brute luck'} that manifest as membership in sensitive attribute groups (since individuals do not choose their gender, ethnicity etc.). The normative principle has been placed within the umbrella of the {\it `justice as fairness'} work due to John Rawls~\cite{rawls1971theory} that underlies most of modern political philosophy. Further, since outlier detection systems are often used to inform human decisions, it is important to ensure that outlier detection algorithms do not propagate and/or reinforce stereotypes present in society by way of placing higher burden on certain sensitive groups than others. 

\section{Background: Local Outlier Factor (LOF)}\label{sec:lof}

Our method builds upon the pioneering LOF framework~\cite{breunig2000lof} for (vanilla) outlier detection. LOF comprises three phases, each computing a value for each object in $\mathcal{X}$, progressively leading to LOF: (i) k-distance, (ii) local reachability density (LRD), and (iii) local outlier factor (LOF). 

\noindent{\bf k-distance:} Let $N_k(X)$ be the set of $k$ nearest neighbors\footnote{$|N_k(X)|$ could be greater than $k$ in case there is a tie for the $k^{th}$ place.} to $X$ (within $\mathcal{X}$), when assessed using the distance function $d(.,.)$. The $k$-distance for each $X \in \mathcal{X}$ is then the distance to the $k^{th}$ nearest object. 

\begin{equation}
    k\text{-}distance(X) = max\{d(X,X') | X' \in N_k(X)\}
\end{equation}

\noindent{\bf Local Reachability Density:} The local reachability density of $X$ is defined as the inverse of the average distance of $X$ to it's $k$ nearest neighbors: 

\begin{equation}
    lrd(X) = 1/ \bigg( \frac{\sum_{X' \in N_k(X)} rd(X,X')}{|N_k(X)|} \bigg)
\end{equation}

where $rd(X,X')$ is an assymetric distance measure that works out to the true distance, except when the true distance is smaller than $k$-$distance(X')$:

\begin{equation}
    rd(X,X') = max \{k\text{-}distance(X'), d(X,X')\}
\end{equation}

This lower bounding by $k$-$distance(X')$ - note also that $k$-$distance(X')$ depends on $N_k(X')$ and not $N_k(X)$ - makes the $lrd(.)$ measure more stable. $lrd(X)$ quantifies the density of the local neighborhood around $X$. 

\noindent{\bf Local Outlier Factor:} The local outlier factor is the ratio of the average $lrd$s of $X$'s neighbors to $X$'s own $lrd$.

\begin{equation}\label{eq:lof}
    lof(X) =  \bigg( \frac{\sum_{X' \in N_k(X)} lrd(X')}{|N_k(X)|} \bigg) / lrd(X)
\end{equation}

An $lof(X) = 1$ indicates that the local density around $X$ is comparable to that of it's neighbors, whereas a $lrd(X) >> 1$ indicates that it's neighbors are in much denser regions than itself. Once $lof(.)$ is computed for each $X \in \mathcal{X}$, the top-$t$ data objects with highest $lof(.)$ scores would be returned as outliers. 

\section{FairLOF: Our Method}


\subsection{Motivation}\label{sec:motivation}


In many cases, the similarity space implicitly defined by the distance function $d(.,.)$ bears influences from the sensitive attributes and grouping of the dataset defined over such attributes. The influence, whether casual, inadvertent or conscious, could cause the sensitive attribute profiles of outliers to be significantly different from the dataset profiles. These could occur in two contrasting ways.

\noindent{\bf Under-reporting of Large/Majority Sensitive Groups:} Consider the case where $d(.,.)$ is aligned with groups defined by $S$. Thus, across the dataset, pairs of objects that share the same value for $S$ are likely to be judged to be more proximal than those that bear different values for $S$. Consider a dataset comprising $75\%$ males and $25\%$ females. Such skew could occur in real-world cases such as datasets sourced from populations in a STEM college or certain professions (e.g., police\footnote{ https://www.statista.com/statistics/382525/share-of-police-officers-in-england-and-wales-gender-rank/}). Let us consider the base/null assumption that {\it real outliers} are also distributed as $75\%$ males and $25\%$ females. Now, consider a male outlier ($M$) and a female outlier ($F$), both of which are equally eligible outliers according to human judgement. First, consider $M$; $M$ is likely to have a {\it quite cohesive} and {\it predominantly male} $k$NN neighborhood due to both: (i) males being more likely in the dataset due to the apriori distribution, and (ii) $d(.,.)$ likely to judge males as more similar to each other (our starting assumption). Note that the first factor works in favour of a male-dominated neighborhood for $F$ too; thus, $F$'s neighborhood would be less gender homogeneous, and thus less cohesive when measured using our $S$ aligned $d(.)$. This would yield $lrd(M) > lrd(F)$, and thus $lof(M) < lof(F)$ (Ref. Eq.~\ref{eq:lof}) despite them being both equally eligible outliers. In short, {\it when $d(.,.)$ is aligned with groups defined over $S$, the smaller groups would tend to be over-represented among the outliers.} 

\noindent{\bf Over-reporting of Large Sensitive Groups:} Consider a domain-tuned distance function designed for a health records agency who would like to ensure that records be not judged similar just due to similarity on gender; such fine-tuning, as is often done with the intent of ensuring fairness, might often be designed with just the {\it 'main groups'} in mind. In the case of gender, this would ensure a good spread of male and female records within the space; however, this could result in minority groups (e.g., LGBTQ) being relegated to a corner of the similarity space. This would result in a tight clustering of records belonging to the minority group, resulting in the LOF framework being unable to pick them out as outliers. Thus, {\it a majority conscious design of $d(.,.)$ would result in over-representation of minority groups among outliers.}


\subsection{FairLOF: The Method}

The construction of {\it FairLOF} attempts to {\it correct} for such $k$NN neighbohood distance disparities across object groups defined over sensitive attributes. {\it FairLOF} distance correction is based on three heuristic principles; (i) {\bf neighborhood diversity} (object-level correction), (ii) {\bf apriori distribution} (value-level), and (iii) {\bf attribute asymmetry} (attribute-level). We outline these for the first scenario in Sec~\ref{sec:motivation}, where $d(.,.)$ is aligned with the sensitive attribute, $S$, resulting in {\it minority over-representation among outliers}; these will be later extended to the analogous scenario, as well as for multiple attributes in $\mathcal{S}$.

\subsubsection{\bf Neighborhood Diversity:} Consider the case of objects that are embedded in neighborhoods comprising objects that take different values of $S$ than itself; we call this as a $S$-diverse neighborhood. These would be disadvantaged with a higher $k$-$distance$, given our assumption that $d(.,.)$ is aligned with $S$. Thus, the $k$-$distance$ of objects with highly diverse neighborhoods would need to be corrected {\it downward}. This is an object-specific correction, with the extent of the correction determined based on $S$-diversity in the object's neighborhood. 

\subsubsection{\bf Apriori Distribution:} Consider objects that belong to an $S$ group that are very much in minority; e.g., LGBTQ groups for $S=gender$. Since these objects would have an extremely diverse neighborhood due to their low apriori distribution in the dataset (there aren't enough objects with the same $S$ value in the dataset), the neighborhood diversity principle would correct them deeply downward. To alleviate this, the neighborhood diversity correction would need to be discounted based on the sparsity of the object's value of $S$ in the dataset.

\subsubsection{\bf Attribute Asymmetry:} The extent of $k$-$distance$ correction required also intuitively depends on the extent to which $d(.,.)$ is aligned with the given $S$. This could be directly estimated based on the extent of minority over-representation among outliers when vanilla {\it LOF} is applied. Accordingly, the attribute asymmetry principle requires that the correction based on the above be amplified or attenuated based on the extent of correction warranted for $S$. 

The above principles lead us to the following form for $k$-$distance$:

\begin{equation}
    \bigg(max\{d(X,X')| X' \in N_k(X)\} \bigg) \bigg( 1 - \lambda \times W^{\mathcal{X}}_S \times D^{\mathcal{X}}_{X.S} \times Div(N_k(X),X.S) \bigg) 
\end{equation}

where $Div(N_k(X),X.S)$, $D^{\mathcal{X}}_{X.S}$ and $W^{\mathcal{X}}_S$ relate to the three principles outlined above (respectively), $\lambda \in [0,1]$ being a weighting factor. These terms are constructed as below:

\begin{align}
    Div(N_k(X),X.S) & = \frac{|\{X'| X' \in N_k(X) \wedge X'.S \neq X.S\}|}{|N_k(X)|} \label{eq:diversity} \\
    D^{\mathcal{X}}_{X.S} & = \frac{|\{X'| X' \in \mathcal{X} \wedge X'.S = X.S\}|}{|\mathcal{X}|} \label{eq:distrib} \\
    W^{\mathcal{X}}_S & = c+ \lvert D^{\mathcal{X}}_{v^*} - D^{\mathcal{R}_{LOF}}_{v^*} \rvert\ \ \text{where} \ \ v^* = \mathop{\arg\max}_{v \in V(S)}\ D^{\mathcal{X}}_v \label{eq:attrweight}
\end{align}

Eq.~\ref{eq:diversity} measures diversity as the fraction of objects among $N_k(X)$ that differ from $X$ on it's $S$ attribute value. Eq.~\ref{eq:distrib} measures the apriori representation as the fraction of objects in $\mathcal{X}$ that share the same $S$ attribute value as that of $X$. For Eq.~\ref{eq:attrweight}, $D^{\mathcal{R}_{LOF}}_v$ refers to the fraction of $S=v$ objects found among the top-$t$ results of vanilla LOF over $\mathcal{X}$. $W^{\mathcal{X}}_S$ is computed as a constant factor (i.e., $c$) added to the asymmetry extent measured as the extent to which the largest $S$-defined group in the dataset is underrepresented in the vanilla LOF results. While we have used a single $S$ attribute so far, observe that this is easily extensible to multiple attributes in $\mathcal{S}$, yielding the following refined form for $k$-$distance$:

\begin{multline}
    \bigg(max\{d(X,X')| X' \in N_k(X)\} \bigg) \bigg( 1 - \lambda \sum_{S \in \mathcal{S}} \big(W^{\mathcal{X}}_S \times D^{\mathcal{X}}_{X.S} \times Div(N_k(X),X.S) \big) \bigg) 
\end{multline}

While we have been assuming the case of $d(.,.)$ aligned with $S$ and minority over-representation among outliers, the opposite may be true for certain attributes in $\mathcal{S}$; recollect the second case discussed in Section~\ref{sec:motivation}. In such cases, the $k$-$distance$ would need to be corrected upward, as against downward. We incorporate that to yield the final $k$-$distance$ formulation for {\it FairLOF}. 

\begin{multline}\label{eq:kdistfair}
    k\text{-}distance_{FairLOF}(X) = \bigg(max\{d(X,X')| X' \in N_k(X)\} \bigg) \times \\
     \bigg( 1 - \lambda \sum_{S \in \mathcal{S}} \big( \mathbb{D}(\mathcal{X},S) \times W^{\mathcal{X}}_S \times D^{\mathcal{X}}_{X.S} \times Div(N_k(X),X.S) \big) \bigg) 
\end{multline}

where $\mathbb{D}(\mathcal{X},S) \in \{-1,+1\}$ denotes the direction of correction as below:

\begin{equation}
    \mathbb{D}(\mathcal{X},S) = \begin{cases}
    +1 & \text{if}\ D^{\mathcal{X}}_{v^*} > D^{\mathcal{R}_{LOF}}_{v^*}\ \ \text{where}\ \ v^* = \mathop{\arg\max}_{v \in V(S)}\ D^{\mathcal{X}}_v \\
    -1 & otherwise
    \end{cases}
\end{equation}

This modification in $k$-$distance$ warrants an analogous correction of $rd(.,.)$ to ensure level ground among the two terms determining $rd(.,.)$.


\begin{multline}\label{eq:rdfair}
    rd_{FairLOF}(X,X') = max \bigg\{ k\text{-}distance_{FairLOF}(X'), \\
     d(X,X') \times \big(1 - \lambda \sum_{S \in \mathcal{S}} \big( \mathbb{D}(\mathcal{X},S) \times W^{\mathcal{X}}_S \times D^{\mathcal{X}}_{X.S} \times \mathbb{I}(X.S \neq X'.S)\} \big) \bigg\}
\end{multline}

The second term in $rd(.,.)$ is corrected in the same manner as for $k$-$distance$, except that the diversity term is replaced by a simple check for inequality, given that there is only one object that $X$ is compared with. 

These distance corrections complete the description of {\it FairLOF}, which is the {\it LOF} framework from Section~\ref{sec:lof} with $k$-$distance(.,.)$ and $rd(.,.)$ replaced by their corrected versions from Eq.~\ref{eq:kdistfair} and Eq.~\ref{eq:rdfair} respectively. While we omit the whole sequence of {\it FairLOF} steps to avoid repetition with Sec.~\ref{sec:lof}, we will use $flof(.)$ to denote the final outlier score from {\it FairLOF}, analogous to $lof(.)$ for LOF. The {\it FairLOF} hyperparameter, $\lambda$, determines the {\it strength} of the fairness correction applied, and could be a very useful tool to navigate the space of options {\it FairLOF} provides, as we will outline in the next section. 

\noindent{\bf Note on Complexity:} Eq.~\ref{eq:distrib} and Eq.~\ref{eq:attrweight} can be pre-computed at a an exceedingly small cost of $\mathcal{O}(|\mathcal{S}| \times m)$ where $m$ is $max_{S \in \mathcal{S}} |V(S)|$. Eq.~\ref{eq:diversity} needs to be computed at a per-object level, thus multiplying the LOF complexity by $|\mathcal{S}| \times m$. With typical values of $|\mathcal{S}| \times m$ being in the 1000s at max (e.g., 3-5 sensitive attributes with 10-20 distinct values each), and outlier detection being typically considered as an offline task not requiring real-time responses, the overheads of the $k$-$distance$ adjustment may be considered as very light. 

\section{Evaluation Framework for Fair Outlier Detection}\label{sec:eval}

Enforcing parity along $S$-groups among outliers, as discussed, often contradicts with identifying high-LOF outliers. This trade-off entails two sets of evaluation measures, inspired by similar settings in fair clustering~\cite{DBLP:conf/edbt/Abraham0S20}. 


\subsubsection{\bf Quality Evaluation:} While the most desirable quality test for any outlier detection framework would be accuracy measured against human generated outlier/non-outlier labels, public datasets with such labels are not available, and far from feasible to generate. Thus, we measure how well {\it FairLOF} results align with the fairness-agnostic {\it LOF}, to assess quality of {\it FairLOF} results. 


\begin{equation}
    Jacc(\mathcal{R}_{LOF}, \mathcal{R}_{FairLOF}) = \frac{|\mathcal{R}_{LOF} \cap \mathcal{R}_{FairLOF}|}{|\mathcal{R}_{LOF} \cup \mathcal{R}_{FairLOF}|}
\end{equation}


\begin{equation}
    Pres(\mathcal{R}_{LOF}, \mathcal{R}_{FairLOF}) = \frac{\sum_{X \in \mathcal{R}_{FairLOF}} lof(X)}{\sum_{X \in \mathcal{R}_{LOF}} lof(X)}
\end{equation}

where $\mathcal{R}_{LOF}$ and $\mathcal{R}_{FairLOF}$ are top-$t$ outliers (for any chosen $k$) from {\it LOF} and {\it FairLOF} respectively. $Jacc(.,.)$ computes the jaccard similarity between the result sets. Even in cases where $\mathcal{R}_{FairLOF}$ diverges from $\mathcal{R}_{LOF}$, we would like to ensure that it does not choose objects with very low $lof(.)$ values within $\mathcal{R}_{FairLOF}$; $Pres(.,.)$ computes the extent to which high $lof(.)$ scores are preserved within $\mathcal{R}_{FairLOF}$, expressed as a fraction of the total $lof(.)$ across $\mathcal{R}_{LOF}$. For both these, higher values indicate better quality of {\it FairLOF} results. 

\subsubsection{\bf Fairness Evaluation:} For any particular sensitive attribute $S \in \mathcal{S}$, we would like the distribution of objects across its values among outliers (i.e., $\mathcal{R}_{FairLOF}$) be similar to that in the dataset, $\mathcal{X}$. In other words, we would like the $\mathcal{D}^{\mathcal{R}_{FairLOF}}_S = [ \ldots, D^{\mathcal{R}_{FairLOF}}_v, \ldots]$ vector ($V \in V(S)$, and Ref. Eq.~\ref{eq:distrib} for computation)  to be as similar as possible to the distribution vector over the dataset for $S$, i.e., $\mathcal{D}^{\mathcal{X}}_S = [ \ldots, D^{\mathcal{X}}_v, \ldots]$, as possible. We would like this to hold across all attributes in $\mathcal{S}$. Note that this fairness notion is very similar to that in fair clustering, the only difference being that we evaluate the outlier set once as against each cluster separately. Thus, we adapt the fairness metrics from~\cite{DBLP:conf/edbt/Abraham0S20,wang2019towards} as below:

\begin{align}
    ED(\mathcal{R}_{FairLOF}) & = \sum_{S \in \mathcal{S}} Euclidean\_Distance(\mathcal{D}^{\mathcal{R}_{FairLOF}}_S, \mathcal{D}^{\mathcal{X}}_S) \\
    Wass(\mathcal{R}_{FairLOF}) & = \sum_{S \in \mathcal{S}} Wasserstein\_Distance(\mathcal{D}^{\mathcal{R}_{FairLOF}}_S, \mathcal{D}^{\mathcal{X}}_S)
\end{align}

where $ED(.)$ and $Wass(.)$ denote aggregated Euclidean and Wasserstein distances across the respective distribution vectors. Since these measure deviations from dataset-level profiles, lower values are desirable in the interest of fairness.


\subsubsection{Quality-Fairness Trade-off:} Note that all the above metrics can be computed without any external labellings. Thus, this provides an opportunity for the user to choose different trade-offs between {\it quality} and {\it fairness} by varying the {\it FairLOF} {\it correction strength} hyper-parameter $\lambda$. We suggest that a practical way of using {\it FairLOF} would be for a user to try with progressively higher values of $\lambda$ from $\{0.1, 0.2, \ldots \}$ (note that $\lambda=0.0$ yields {\it FairLOF} $=$ {\it LOF}, with higher values reducing $Jacc(.)$ and $Pres(.)$ progressively) using a desired value of {\it Jaccard} similarity as a pilot point. For example, we may want to retain a Jaccard similarity (i.e., $Jacc(.)$ value) of approximately $0.9$ or $0.8$ with the original {\it LOF} results. Thus, the user may stop when that is achieved. The quantum of fairness improvements achieved by {\it FairLOF} over {\it LOF} at such chosen points will then be indicative of {\it FairLOF}'s effectiveness.




\subsubsection{Single Sensitive Attribute and a Quota-based System:} As noted upfront, {\it FairLOF} is targeted towards cases where there are multiple sensitive attributes to ensure fairness over; this is usually the case since there are often many sensitive attributes in real world scenarios (e.g., gender, ethnicity, caste, religion, native language, marital status, and even age in certain scenarios). For the case of a single sensitive attribute with a handful of possible values, there is a simple and effective strategy for fair outlier detection. Consider $S = gender$; we take the global list of objects sorted in descending order of $lof(.)$ scores and splice them into {\it male} list, {\it female} list etc. Based on the desired distribution of gender groups among outliers (as estimated from the dataset), {\it `quotas'} may be set for each gender value, and the appropriate number of top objects from each gender-specific list is then put together to form the outlier set of $t$ objects. This is similar to the strategy used for job selection in India's affirmative action policy ({\it aka} reservation\footnote{https://en.wikipedia.org/wiki/Reservation\_in\_India}). The extension of this quota-based strategy to multiple sensitive attributes by modelling them as one giant attribute taking values from the cross-product, is impractical due to multiple reasons. First, the cross-product may easily exceed $t$, leading to practical and legal issues across scenarios; for example, with just $\mathcal{S} = \{nationality, ethnicity\}$, we could have the cross product approaching $2000+$ given there are $200+$ nationalities and at least $10+$ ethnicities, and practical values of $t$ could be in the 100s due to manual perusal considerations. Thus, the quota system, by design, would exclude the large majority of rare combinations of sensitive attribute values from being represented among outliers however high their LOF scores may be; such a policy is unlikely to survive any legal or ethical scrutiny to allow practical uptake. Second, the quota based system offers no way to control the trade-off between fairness and quality, making it impractical to carefully choose trade-offs as outlined earlier. Third, even a simpler version of the extension of the quota-based system to multiple attributes has been recently shown to be NP-hard~\cite{nphardaamas2020}.

\begin{table}[h!]
\begin{center}
\begin{tabular}{|l|l|l|l|}
\hline
{\bf Dataset} & {\bf Domain} & {\bf $|\mathcal{X}|$} & {\bf Sensitive Attributes Used} \\
\hline
Adult\tablefootnote{http://archive.ics.uci.edu/ml/datasets/Adult} & US 1994 Census & 48842 & {\it marital status, race, sex, nationality} \\
\hline
CC\tablefootnote{https://archive.ics.uci.edu/ml/datasets/default+of+credit+card+clients} & Credit Card Default & 30000 & {\it sex, education, marital status} \\
\hline
W4HE\tablefootnote{https://archive.ics.uci.edu/ml/datasets/wiki4he} & Wikipedia HE Use & 913 & {\it gender, disciplinary domain, uni name} \\
\hline
St-Mat\tablefootnote{https://archive.ics.uci.edu/ml/datasets/Student+Performance} & Student Maths Records & 649 & {\it gender, age} \\
\hline
\end{tabular}
\caption{Dataset Information}
\label{tab:data}
\end{center}
\end{table}

\section{Experimental Evaluation}

\noindent{\bf Datasets and Experimental Setup:} There are only a few public datasets with information of people, the scenario that is most pertinent for fairness analysis; this is likely due to person-data being regarded highly personal and anonymization could still lead to leakage of identifiable information\footnote{https://en.wikipedia.org/wiki/AOL\_search\_data\_leak}. The datasets we use along with details are included in Table~\ref{tab:data}. The datasets encompass a wide variety of scenarios, and vary much in sizes as well as the sensitive attributes used. We set $t$ (to get top-$t$ results) to $5\%$ of the dataset size capped at $500$, $k=5$ and $c$ (Ref. Eq.~\ref{eq:attrweight}) to be $1/s$ where $s$ is the number of sensitive attributes. 

\begin{table}[h!]
\begin{center}
\begin{tabular}{|l|l||p{1.1cm}p{1.1cm}||p{1.1cm}p{1.1cm}||p{1.1cm}p{1.1cm}||p{1.1cm}p{1.1cm}||}
\hline
\hline
{\bf Dataset} & {\bf Guide} & \multicolumn{4}{c||}{{\bf Quality}} & \multicolumn{4}{c||}{{\bf Fairness}} \\
\cline{3-10}
& {\bf Point} & {\it Jacc} & {\it Det\%} & {\it Pres} & {\it Det\%} & {\it ED} & {\it Impr\%} & {\it Wass} & {\it Impr\%} \\
\hline
\hline
\multirow{3}{*}{{\bf Adult}} & LOF & 1.0 & N/A & 1.0 & N/A & 0.2877 & N/A & 0.5328 & N/A \\
\cline{2-10}
& 0.9 & 0.8939 & 10.61\% & 0.9977 & {\bf 00.23\%} & 0.1906 & {\bf 33.75\%} & 0.3468 & {\bf 34.91\%} \\
& 0.8 & 0.7986 & 20.14\% & 0.9906 & {\bf 00.94\%} & 0.1714 & {\bf 40.42\%} & 0.2372 & {\bf 55.48\%} \\
\hline
\hline
\multirow{3}{*}{{\bf CC}} & LOF & 1.0 & N/A & 1.0 & N/A & 0.2670 & N/A & 0.2152 & N/A \\
\cline{2-10}
& 0.9 & 0.9011 & 09.89\% & 0.9976 & {\bf 00.24\%} & 0.2235 & 16.29\% & 0.2112 & 01.86\% \\
& 0.8 & 0.7921 & 20.79\% & 0.9879 & 01.21\% & 0.1568 & {\bf 41.27\%} & 0.2012 & 06.51\% \\
\hline
\hline
\multirow{3}{*}{{\bf W4HE}} & LOF & 1.0 & N/A & 1.0 & N/A & 0.2121 & N/A & 0.3305 & N/A \\
\cline{2-10}
& 0.9 & 0.8776 & 12.24\% & 0.9987 & {\bf 00.13\%} & 0.0966 & {\bf 54.46\%} & 0.2989 & 09.56\% \\
& 0.8 & 0.8039 & 19.61\% & 0.9951 & {\bf 00.49\%} & 0.1820 & 14.19\% & 0.2498 & {\it 24.42\%} \\
\hline
\hline
\multirow{3}{*}{{\bf St-Mat}} & LOF & 1.0 & N/A & 1.0 & N/A & 0.4174 & N/A & 0.9196 & N/A \\
\cline{2-10}
& 0.9 & 0.9047 & 09.53\% & 0.9970 & {\bf 00.30\%} & 0.3467 & 16.94\% & 0.8962 & 02.54\% \\
& 0.8 & 0.8182 & 18.18\% & 0.9896 & 01.04\% & 0.3467 & 16.94\% & 0.8962 & 02.54\% \\
\hline
\hline
\end{tabular}
\caption{FairLOF Effectiveness Study. The {\it FairLOF} results at {\it guide points} set to $Jacc=0.9$ and $Jacc=0.8$ are shown along with LOF results for each of the datasets. Since we use coarse steps for $\lambda$, the precise guide point value for $Jacc$ may not be achieved; so we choose the closest $Jacc$ that is achievable to the guide point. The deteriorations in Quality metrics and improvements in Fairness metrics are indicated in percentages. Fairness improvements of $20\%+$ are italicized, and those that are $30\%+$ are shown in bold, whereas Quality deteriorations $<1\%$ are indicated in bold.}
\label{tab:effectiveness}
\end{center}
\vspace{-0.3in}
\end{table}

\subsection{FairLOF Effectiveness Study}

The effectiveness of {\it FairLOF} may be assessed by considering the quantum of fairness achieved at low degradations to quality. {\it It may be noted that higher values are better on the quality measures ({\it Jacc} and {\it Pres}) and lower values are better on the fairness measures ({\it ED} and {\it Wass})}. We follow the quality-fairness trade-off strategy as outlined in Section~\ref{sec:eval} with a $\lambda$ search step-size of $0.1$ and choose $0.9$ and $0.8$ as {\it guide points} for $Jacc$. The detailed results are presented in Table~\ref{tab:effectiveness}, with details of the result formatting outlined in the caption therein. Broadly, we observe the following:

\begin{itemize}[leftmargin=*]
    \item {\bf Fairness Improvements:} {\it FairLOF} is seen to achieve significant improvements in fairness metrics at reasonable degradations to quality. The $ED$ measure is being improved by $30\%$ on an average at the chosen guide points, whereas $Wass$ is improved by $12\%$ and $22\%$ on an average at the guide points of $0.9$ and $0.8$ respectively. These are evidently hugely significant gains indicating that {\it FairLOF} achieves compelling fairness improvements. 
    \item {\bf Trends on Pres:} Even at $Jacc$ close to $0.9$ and $0.8$, the values of $Pres$ achieved by {\it FairLOF} are seen to be only marginally lower than $1.0$, recording degradations of less than $1.0\%$ in the majority of the cases. This indicates that while the {\it LOF} results are being altered, {\it FairLOF} is being able to replace them with other objects with substantively similar $lof(.)$ values. This, we believe, is a highly consequential result, indicating that {\it FairLOF} remains very close in spirit to {\it LOF} on result quality while achieving the substantive fairness gains. 
\end{itemize}

In addition to the above, we note the following trends on {\it FairLOF} performance. First, $Wass$ is significantly harder to optimize for, as compared to $ED$; this is because $Wass$ prefers the gains to be equally distributed across sensitive attributes. Second, for small datasets where there is relatively less room for re-engineering outlier results for fairness, {\it FairLOF} gains are seen to saturate quickly. This is most evident for {\it St-Mat} in Table~\ref{tab:effectiveness}. 

\subsection{FairLOF Parameter Sensitivity Study}

One of the key aspects is to see whether {\it FairLOF} effectiveness is smooth against changes in $\lambda$, the only parameter of significant consequence in {\it FairLOF}. In particular, we desire to see consistent decreases on each of {\it Jacc}, {\it Pres}, {\it ED} and {\it Wass} with increasing $\lambda$. On each of the datasets, such gradual and smooth trends were observed, with the gains tapering off sooner in the case of the smaller datasets, W4HE and St-Mat. The trends on Adult and CC were very similar; for Adult, we observed that the Pearson product-moment correlation co-efficient~\cite{pearson1895vii} against $\lambda \in [0,1]$ to be -0.900 for {\it Jacc}, -0.973 for {\it Pres}, -0.997 for {\it ED} and -0.959 for {\it Wass} indicating a graceful movement along the various metrics with changing $\lambda$. We observed similar consistent trends for increasing $c$ (Eq.~\ref{eq:attrweight}) as well. {\it FairLOF} was also observed to be quite stable with changes of $k$ and $t$.

\section{Conclusions and Future Work}

In this paper, for the first time (to our best knowledge), we considered the task of fair outlier detection. Fairness is of immense importance in this day and age when data analytics in general, and outlier detection in particular, is being used to make and influence decisions that will affect human lives to a significant extent, especially within web data scenarios that operate at scale. We consider the paradigm of local neighborhood based outlier detection, arguably the most popular paradigm in outlier detection literature. We outlined the task of fair outlier detection over a plurality of sensitive attributes, basing our argument on the normative notion of luck egalitarianism, that the costs of outlier detection be borne proportionally across groups defined on protected/sensitive attributes such as gender, race, religion and nationality. We observed that using a task-defined distance function for outlier detection could induce unfairness when the distance function is not fully orthogonal to all the sensitive attributes in the dataset. We develop an outlier detection method, called {\it FairLOF}, inspired by the construction of {\it LOF} and makes use of three principles to nudge the outlier detection towards directions of increased fairness. We outline an evaluation framework for fair outlier detection, and use that in evaluating {\it FairLOF} extensively over real-world datasets. Through our empirical results, we observe that {\it FairLOF} is able to deliver substantively improved fairness in outlier detection results, at reasonable detriment to result quality as assessed against {\it LOF}. This illustrates the effectiveness of {\it FairLOF} in achieving fairness in outlier detection. 

\noindent{\bf Future Work:} In this work, we have limited our attention to local neighborhood based outlier detection. Extending notions of fairness to global outlier detection would be an interesting future work. Further, we are considering extending {\it FairLOF} to the related task of identifying groups of anomalous points, and other considerations of relevance to fair unsupervised learning~\cite{whitherfc}. 


\bibliographystyle{splncs04}
\bibliography{refs}

\begin{thebibliography}{10}
\providecommand{\url}[1]{\texttt{#1}}
\providecommand{\urlprefix}{URL }
\providecommand{\doi}[1]{https://doi.org/#1}

\bibitem{DBLP:conf/edbt/Abraham0S20}
Abraham, S.S., P, D., Sundaram, S.S.: Fairness in clustering with multiple
  sensitive attributes. In: EDBT. pp. 287--298 (2020)

\bibitem{asudeh2019designing}
Asudeh, A., Jagadish, H., Stoyanovich, J., Das, G.: Designing fair ranking
  schemes. In: SIGMOD (2019)

\bibitem{babaei2019detecting}
Babaei, K., Chen, Z., Maul, T.: Detecting point outliers using prune-based
  outlier factor (plof). arXiv preprint arXiv:1911.01654  (2019)

\bibitem{barocas2016big}
Barocas, S., Selbst, A.D.: Big data's disparate impact. Calif. L. Rev.
  \textbf{104}, ~671 (2016)

\bibitem{nphardaamas2020}
Bei, X., Liu, S., Poon, C.K., Wang, H.: Candidate selections with proportional
  fairness constraints. In: AAMAS (2020)

\bibitem{DBLP:conf/nips/BeraCFN19}
Bera, S.K., Chakrabarty, D., Flores, N., Negahbani, M.: Fair algorithms for
  clustering. In: NeurIPS. pp. 4955--4966 (2019)

\bibitem{breunig2000lof}
Breunig, M.M., Kriegel, H.P., Ng, R.T., Sander, J.: Lof: identifying
  density-based local outliers. In: SIGMOD. pp. 93--104 (2000)

\bibitem{chandola2007outlier}
Chandola, V., Banerjee, A., Kumar, V.: Outlier detection: A survey. ACM
  Computing Surveys  \textbf{14}, ~15 (2007)

\bibitem{chawla2006slom}
Chawla, S., Sun, P.: Slom: a new measure for local spatial outliers. Knowledge
  and Information Systems  \textbf{9}(4),  412--429 (2006)

\bibitem{chen2017outlier}
Chen, J., Sathe, S., Aggarwal, C., Turaga, D.: Outlier detection with
  autoencoder ensembles. In: SDM (2017)

\bibitem{chierichetti2017fair}
Chierichetti, F., Kumar, R., Lattanzi, S., Vassilvitskii, S.: Fair clustering
  through fairlets. In: NIPS (2017)

\bibitem{10.1145/3376898}
Chouldechova, A., Roth, A.: A snapshot of the frontiers of fairness in machine
  learning. Commun. ACM  \textbf{63}(5),  82–89 (Apr 2020)

\bibitem{fairod2020}
Davidson, I., Ravi, S.: A framework for determining the fairness of outlier
  detection. In: ECAI (2020)

\bibitem{domingues2018comparative}
Domingues, R., Filippone, M., Michiardi, P., Zouaoui, J.: A comparative
  evaluation of outlier detection algorithms: Experiments and analyses. Pattern
  Recognition  \textbf{74},  406--421 (2018)

\bibitem{fan2011unsupervised}
Fan, W., Bouguila, N., Ziou, D.: Unsupervised anomaly intrusion detection via
  localized bayesian feature selection. In: ICDM (2011)

\bibitem{hawkins1980identification}
Hawkins, D.M.: Identification of outliers, vol.~11. Springer (1980)

\bibitem{huang2019stable}
Huang, L., Vishnoi, N.K.: Stable and fair classification. arXiv:1902.07823
  (2019)

\bibitem{jabez2015intrusion}
Jabez, J., Muthukumar, B.: Intrusion detection system (ids): anomaly detection
  using outlier detection approach. Procedia Computer Science  \textbf{48},
  338--346 (2015)

\bibitem{knight2009luck}
Knight, C.: Luck egalitarianism: Equality, responsibility, and justice. EUP
  (2009)

\bibitem{kriegel2009loop}
Kriegel, H.P., Kr{\"o}ger, P., Schubert, E., Zimek, A.: Loop: local outlier
  probabilities. In: CIKM (2009)

\bibitem{kumar2008outlier}
Kumar, V., Kumar, D., Singh, R.: Outlier mining in medical databases: an
  application of data mining in health care management to detect abnormal
  values presented in medical databases. IJCSNS International Journal of
  Computer Science and Network Security pp. 272--277 (2008)

\bibitem{olfat2019convex}
Olfat, M., Aswani, A.: Convex formulations for fair principal component
  analysis. In: AAAI. vol.~33, pp. 663--670 (2019)

\bibitem{whitherfc}
P, D.: Whither fair clustering? In: AI for Social Good Workshop (2020)

\bibitem{patro2019incremental}
Patro, G.K., et~al.: Incremental fairness in two-sided market platforms: On
  updating recommendations fairly. In: AAAI (2020)

\bibitem{pawar2014survey}
Pawar, A.D., Kalavadekar, P.N., Tambe, S.N.: A survey on outlier detection
  techniques for credit card fraud detection. IOSR J. Comp. Engg.
  \textbf{16}(2),  44--48 (2014)

\bibitem{pearson1895vii}
Pearson, K.: Vii. note on regression and inheritance in the case of two
  parents. proceedings of the royal society of London  \textbf{58}(347-352),
  240--242 (1895)

\bibitem{rawls1971theory}
Rawls, J.: A theory of justice. Harvard university press (1971)

\bibitem{schubert2014local}
Schubert, E., Zimek, A., Kriegel, H.P.: Local outlier detection reconsidered: a
  generalized view on locality with applications to spatial, video, and network
  outlier detection. Data Mining and Knowledge Discovery  \textbf{28}(1),
  190--237 (2014)

\bibitem{wang2019towards}
Wang, B., Davidson, I.: Towards fair deep clustering with multi-state protected
  variables. arXiv preprint arXiv:1901.10053  (2019)

\bibitem{yu2002findout}
Yu, D., Sheikholeslami, G., Zhang, A.: Findout: Finding outliers in very large
  datasets. Knowledge and Information Systems  \textbf{4}(4),  387--412 (2002)

\bibitem{zafar2015fairness}
Zafar, M.B., Valera, I., Rodriguez, M.G., Gummadi, K.P.: Fairness constraints:
  Mechanisms for fair classification. arXiv preprint arXiv:1507.05259  (2015)

\bibitem{zehlike2017fa}
Zehlike, M., Bonchi, F., Castillo, C., Hajian, S., Megahed, M., Baeza-Yates,
  R.: Fa* ir: A fair top-k ranking algorithm. In: CIKM. pp. 1569--1578 (2017)

\bibitem{zhang2009new}
Zhang, K., Hutter, M., Jin, H.: A new local distance-based outlier detection
  approach for scattered real-world data. In: PAKDD (2009)

\end{thebibliography}

\end{document}